\documentclass[letterpaper]{article} 
\usepackage{aaai23}  
\usepackage{times}  
\usepackage{helvet}  
\usepackage{courier}  
\usepackage[hyphens]{url}  
\usepackage{graphicx} 
\usepackage{natbib}  
\usepackage{caption} 
\usepackage{mathtools,halloweenmath}
\usepackage{algorithm}
\usepackage{multirow}
\usepackage[noend]{algpseudocode}
\usepackage{newfloat}
\usepackage{listings}
\usepackage{verbatim} 
\DeclareCaptionStyle{ruled}{labelfont=normalfont,labelsep=colon,strut=off} 
\floatstyle{ruled}
\newfloat{listing}{tb}{lst}{}
\floatname{listing}{Listing}

\usepackage{booktabs}
\usepackage{tabularx} 
\usepackage{xcolor}

\nocopyright

\title{Neural Story Planning}
\author{
    Anbang Ye, Christopher Cui, Taiwei Shi, and Mark O. Riedl}
\affiliations{
    Georgia Institute of Technology\\

    \{aye42, ccui46, maksimstw, riedl\}@gatech.edu

}

\begin{document}
\maketitle

\begin{abstract}
Automated plot generation is the challenge of generating a
sequence of events that will be perceived by readers as the
plot of a coherent story.
Traditional symbolic planners plan a story from a goal state and guarantee logical causal plot coherence but rely on a library of hand-crafted actions with their preconditions and effects. 
This closed world setting limits the length and diversity of what symbolic planners can generate. On the other hand, pre-trained neural language models 
can generate stories with great diversity, while being generally incapable of ending a story in a specified manner and can have trouble maintaining coherence.
In this paper, we present an
approach to story plot generation that unifies causal planning with neural language models.
We propose to use commonsense knowledge extracted from large language models to recursively expand a story plot in a backward chaining fashion. 
Specifically, our system infers the preconditions for events in the story and then events that will cause those conditions to become true.
We performed automatic evaluation to measure narrative coherence as indicated by the ability to answer questions about whether different events in the story are causally related to other events. 
Results indicate that our proposed method produces more coherent plotlines than several strong baselines.
\end{abstract}
\section{Introduction}

Automated plot generation is the challenge of generating a sequence of events that will be perceived by readers as the plot of a coherent story. Early solutions to story generation included symbolic planning \cite{Meehan1977TALESPINAI,Lebowitz1985StorytellingAP,Porteous2009ControllingNG,Riedl_2010,CPOCL}, and case-based reasoning~\cite{MEXICA, 10.1016/j.knosys.2004.10.011}. 
These techniques explicitly represent and reason about the causal relationship between events.
In particular, symbolic planners infer {\em causal relations} between events and
through these causal relations, the logical soundness of a plan could be guaranteed.
Story planners extended generic symbolic planners in a number of ways to create character believability~\cite{Riedl_2010}, character conflict~\cite{CPOCL}, theory of mind~\cite{Ware_Siler_2021}, etc. 
In story generators, logical soundness corresponds with the concept of {\em plot coherence} where each event---and the goal state---is justified by preceding events or the initial world state.
Narrative psychologists note the importance of plot coherence in reader comprehension, avoiding confusion, acceptance of stories, and memory of stories~\cite{TRABASSO1985612,Graesser1991QuestionAI}.

Unfortunately provable plot coherence comes at a cost.
Symbolic planners require libraries of hand-crafted schemas that describe what actions (called events when planning stories) are available.
Among other things, action schemas define {\em preconditions}---logical statements that must be true for an event to be executable---and {\em effects}---logical statements that describe how the state of the world is changed if an action succeeds in execution. 
While planned stories are highly causally coherent, the reliance on hand-crafted libraries of action schemas 
and 
pre-existing symbols for characters, objects, and locations, limits the length and diversity of plots.

Neural language model based approaches to story generation, on the other hand, can generate more diverse stories about any number of topics included in the training data~\cite{Martin,roemmele-gordon-2018-encoder, fan-etal-2018-hierarchical,Yao_Peng_Weischedel_Knight_Zhao_Yan_2019,plan-write-revise,DBLP:journals/corr/abs-2004-14967}.
This is especially true for large, pre-trained neural language models
that have been trained on, amongst other things, large corpora of stories.
Neural story generators create stories by sampling from the probability distribution $P(w_n|w_1, w_2, ..., w_{n-1};\theta)$ where $\theta$ are the parameters of the language model.
Sampling from this learned distribution emulates human patterns of language, including that found in stories.
However, when addressing novel combinations of topics and circumstances sampling is not guaranteed to provide causal coherence.
Other limitations of neural text generation, such as repetition and generic responses, have also been documented~\cite{holtzman-degeneration}.

In this paper, we seek to unify the strengths of symbolic planning and neural text generation to achieve causally coherent, ending-guided, story plotlines.
To this end, we have developed a story planner based on {\em partial-order causal link planning} (POCL)~\cite{Penberthy1992UCPOPAS} to generate story plotlines.
However, instead of using a library of hand-crafted action schemas and world state symbols, our planner uses a large pre-trained language model (specifically GPT-J-6B) to infer events and their preconditions.
In this way, the story planner is able to operate without pre-specifiying actions, characters, world locations, or objects in the world. 
Yet the planner can still identify causal relations between events and ensure causal coherence of stories.

\begin{figure*}
\includegraphics[width=\textwidth]{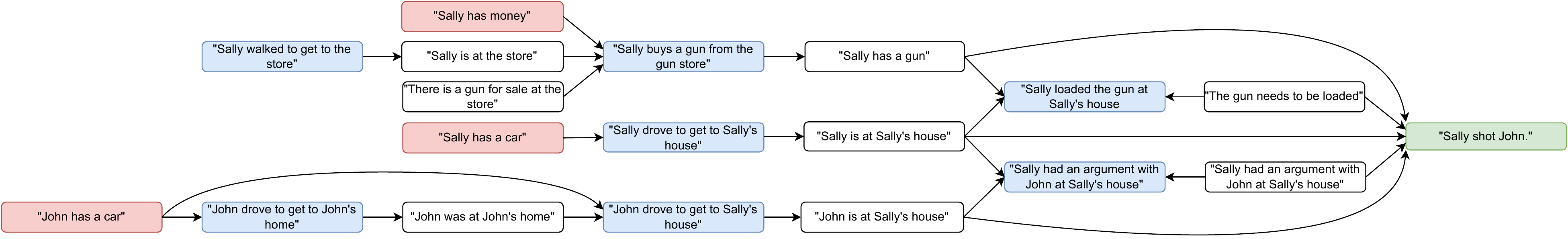}
  \caption{Visualization of a partial order plan for a plotline. The final event, which is given as input, is green.
  Events are blue. 
  Preconditions are white and red; red preconditions are those that matched against given initial conditions.
  Arcs from events to preconditions are {\em causal links}, indicating that the event makes the condition true.
  Arcs from precondition to event indicate that the former is a necessary condition of the event. 
  The plan is generated from right to left, starting with the final event.
  The plan can be totally-ordered, as shown in Table~\ref{tab:topological}.
  }

\label{fig:plan-graph}
\end{figure*}

Our planner works via ends-means chaining---working backward from a given story ending. 
Given a text description of the ending of the story as well as an optinal set of initial-state conditions that describe elements of the story world that can be assumed true, the system queries the language model to generate preconditions that must be satisfied for the given ending to occur.
For example (see Figure~\ref{fig:plan-graph}), 
the event, ``Sally buys a gun from the gun store'' will produce preconditions that include ``Sally is at the story'' and ``Sally has money''.
Each of these preconditions describe a partial world state that must come into existence through the effects of preceding events. 
Next, the system queries the language model for an event that would bring about each precondition. 
Continuing the example, ``Sally is at the store'' produces a new event: ``Sally walked to get to the store''.
The process repeats,
resulting in a graph---a partially-ordered plan of events---that can be topologically flattened into the plot line of a story.

Our system outperforms other strong baselines in generating coherent plotlines as determined by ability to answer questions about the causal relations between events.
The ability to answer questions about stories is an indicator of support for reader comprehension~\cite{Graesser1991QuestionAI}.

\section{Background and Related Work}

Formally, a planner finds a sequence of actions that transforms the initial world state into one in which a goal situations holds.
Story and plot generation using symbolic planning~\cite{Meehan1977TALESPINAI, Lebowitz1985StorytellingAP,Porteous2009ControllingNG,Riedl_2010,CPOCL, Ware_Siler_2021} is known to result in causally coherent plotlines but at the expense of story diversity and length due to reliance on hand-crafted action schemas and  symbols. 

\subsection{{Partial Order Causal Link Planning}}

Several story planners~\cite{Riedl_2010,CPOCL} use a form of symbolic planning called {\em partial order causal link planning} (POCL)~\cite{Penberthy1992UCPOPAS}.
POCL planners search plan-space where every node in the space constitutes an unique partially-ordered plan and the planner transitions from one plan to the next by adding an action or resolving a logical causal flaw due to partial ordering.
A plan is represented as a tuple $P=\langle A,C,O\rangle$. $A$ is a set of actions instantiated from a library of action schema templates that specify preconditions effects.
$C$ is a set of {\em causal links} of the form $a_1 \xrightarrow{c} a_2$ where $a_1,a_2\in A$ and $c$ is a predicate condition that unifies with an effect of $a_1$ and a precondition of $a_2$.
$O$ is a set of temporal ordering constraints $a_1\rightarrow a_2$ that indicates that $a_1$ must be temporally ordered before $a_2$.

At every iteration, the planner selects a plan on the fringe of the plan-space and an action in that plan with an unsatisfied precondition (or the goal state).
An unsatisfied precondition is one in which there does not exist a causal link that points to the action that has a matching the condition. 
A new successor plan is generated for each way of satisfying the action precondition---either by adding a new action and causal link, or by identifying an existing, preceding action with a matching effect (or the initial state)---and extending a causal link between them.
Other operations not summarized here ensure logical soundness.
Planning terminates when a plan is found with no actions with unsatisfied preconditions. 

While there are newer, more computationally efficient symbolic planners, we start with POCL planners for three reasons. 
First, they explicitly identify preconditions that must be satisfied and explicitly identify actions that satisfy preconditions, making plans explainable.
Second, the explicit representation of actions and preconditions makes it amenable to adaptation to the use of LM inferences.
Extending our proposed approach to more modern planning algorithms is left for future extensions of this work.
Third, POCL planners are cognitively plausible ends-means planning processes for humans~\cite{young1999notes}.

\subsection{Neural Story Generators}

Early attempts at neural story generation used language models trained on story corpora to generate stories~\cite{roemmele-gordon-2018-encoder} or plots~\cite{Martin}. 
However, such vanilla use of language models were found to have trouble maintaining coherent context.
\citet{CSKESG} further fine-tunes a pre-trained language model on commonsense datasets to help increase the coherence of stories.
Attempts to control the coherence of stories and plots include conditioning~(e.g., \citet{DBLP:journals/corr/abs-2004-14967}) or hierarchical generation~\cite{fan-etal-2018-hierarchical,Yao_Peng_Weischedel_Knight_Zhao_Yan_2019, Fan2019}.
Hierarchical generation is similar to greedy planning by first generating a ``sketch'' or ``skeleton'' at a higher level of abstraction.

Language models are not generally aware of, or able to drive toward, a given story ending.
\citet{Tambwekar2018ControllableNS} fine-tuned language models to be goal-aware. The EDGAR system~\cite{storyGen_via_QA} generates stories backward using language model based question-answering to iteratively ask how the story state came to be.
This is similar to our approach, but does not explicitly reason about causality.

C2PO~\cite{Ammanabrolu2021AutomatedSV} conducts a bi-directional search, querying story events that can possibly follow or possibly precede events using the COMET~\cite{Bosselut2019COMETCT} model of commonsense inference.
C2PO is the closest to our approach in terms of conducting a search through event space.
C2PO relations between events are referred to as {\em soft} causal links because the capture inferred (probabilistic) event relationships, whereas our system uses {\em hard} causal links that capture explicit conditional relations between events as in POCL planning.

The TattleTale system~\cite{simon-spark-2022} uses a classical symbolic planner to generate a totally ordered sequence of actions and then uses this sequence to condition a language model to generate natural language stories.
The symbolic planner uses hand-crafted action schemas.

\section{Neural Story Planner}

Our proposed technique generates causally coherent story plans using the concepts of causal links from POCL planning. 
However, unlike POCL planning, events and their preconditions are inferred by a large language model (LM) and represented as text descriptions.
An example plan generated by our system is shown in Figure~\ref{fig:plan-graph} with corresponding total-ordered plot in Table~\ref{tab:topological}.

Action preconditions can be thought of as commonsense rules for action applicability.
Commonsense knowledge refers to commonly held beliefs about our physical world.
Because our planner works in an open world setting we must infer what preconditions need to be satisfied to enable an action, and what actions can be taken to satisfy those preconditions.
Language models above 1B parameters have been demonstrated to capably answer commonsense questions~\cite{Bisk2020} and 
we use the GPT-J-6B\footnote{\url{https://github.com/kingoflolz/mesh-transformer-jax}} language model to infer both events and preconditions.

\begin{table}[t]
\centering
\begin{tabular}{l}
\hline
{\bf Ending: Sally shot John}\\ 
\hline
Sally walked to get to store.\\ 
Sally buy a gun from the store.\\
Sally drove to get to Sally's house.\\ 
Sally loaded the gun.\\ 
John drove to get to John's home.\\ 
John drove to get to Sally's house.\\ 
Sally had an argument with John at Sally's house.\\ 
Sally shoot John.\\
\hline
\end{tabular}
\caption{An example of ending-guided planning using neural planner. Events from the plan in Figure~\ref{fig:plan-graph} are shown in chronological order.}
\label{tab:topological}
\end{table}

Our planner starts with an input sentence $s$ that specifies the ending of a story. 
The planner is also provided with a set of initial conditions $I$, which are sentences describing facts  known to be true about the world before the story begins.
The system then recursively uses the LM to infer preconditions of any event in the plot, beginning with $s$ and, for each precondition, uses the LM again to infer an event that would cause that condition to become true. 
See Algorithm~\ref{alg:pseudoPSO}.

The algorithm is a variation on POCL planning, with a few modifications. 
For any given precondition, we semi-greedily accept the first event inferred by the LM. 
The open-world setting is such that the logical impossibility that happens in closed worlds with fixed symbols rarely happens and we find the exhaustic plan-space expansion to be unnecessary.
However, some inferences result in cycles of repeated events, in which case the planner backtracks and selects alternative inferences.

\subsection{Precondition Generation}

We use the few-shot capabilities of large language models to infer the preconditions of events. 
We designed prompts based on psychological theories of reading comprehension that indicate that readers actively track different types of relations between events~\cite{TRABASSO1985612,Graesser1991QuestionAI}. 
\textit{Consequence} relations are equivalent to causal links and indicate that the latter event cannot occur unless the former occurs.
Thus consequence relations also capture whether one event enables another to be possible.
\textit{Reason} relations capture the goals of characters that explain why they are doing something.
\textit{Outcome} relations capture which events satisfy which goals.
\textit{Initiate} relations capture when an event causes a character to form a goal.

We break preconditions into six different classes
and have developed a prompt for each. 
The first four classes capture {\em consequence} relations:
\begin{itemize}
\item {\bf Item Need.} What item(s) must the character possess for an action to execute?
E.g., in order for one character to drive somewhere they must have a car.
\begin{quote}
What item must $[Person]$ possess to $[Action]$?\\Answer: $[Person]$ must have $[Answer]$
\end{quote}

    \item {\bf Location.} Where must the character be in order to carry out an action?
    E.g., for one character to buy something from a store the character must be in the store.
    \begin{quote}
        $[Context]$\\ 
        $[Event]$.\\
        $[Optional~Hint]$\\
        Where is $[Person]$? : $[Answer]$ \\
    \end{quote}
    \vspace{-1\baselineskip}
    The context consists of all events that are ancestors of the current event; 
    even though they will come later in the story, they bias the LM toward reusing locations.
    The hint, which is optional, is filled with any location information that is heuristically extracted from the event.

    \item {\bf Item state.} What must be true about an item for an action to be performed?
    E.g., in order to shoot someone, the gun must be loaded.
    \begin{quote}
        $[Event]$\\
        What can we tell about the $[item]$ other than how $[Person]$ obtained it? $[Answer]$
    \end{quote}

    \item {\bf How.} What needed to have happened for something to be achievable now?
    E.g., ``John became a pro wrestler'' would have a how-precondition that could be satisfied by another event such as ``John trained for one year''. 
    \begin{quote}
        $[Context]$\\
        $[Event]$\\
        How does $[Event]$? $[Answer]$
    \end{quote}
\end{itemize}
The next precondition class aligns with {\em initiates} relations.
Our version focuses on character-character interactions.
\begin{itemize}
    \item {\bf Interactions with others.} What must have happened between two characters for an action to be peformed?
    For example, a character might perform a violent action toward another character if first there was an argument.
    \begin{quote}
        $[Context]$\\
        $[Event]$\\
        Prior to $[Event]$, $[Answer]$
    \end{quote}
\end{itemize}
The final class of precondition is the {\em reason}, which can be satisfied by an event that causes a character to have a goal:
\begin{itemize}
    \item {\bf Reason.} What serves as the reason for an event?
    E.g., the event ``John was arrested'' would have a reason-precondition that could be satisfied by an event such as ``John stole a car''.
    \begin{quote}
        $[Context]$
        $[Event]$ because $[Answer]$    
    \end{quote}
\end{itemize}
Each prompt consists of seven few-shot examples and the 8th leaves the answer blank to be completed by the LM.

When the inference is a short phrase, as is the case in all of our precondition classes, but especially true for {\em Item Need} preconditions, querying a LM suffers from surface form competition \cite{holtzman-etal-2021-surface} in which 
very common responses can be favored over those that are more responsive to the prompt. 
For example, to the question ``What does John need to clean the table?" a LM would prefer the answer ``nothing" over the correct answer ``cleaning cloth" because GPT-J has seen the former more often than the latter in its training corpus.
We adapt the {\em Domain Conditional Pointwise Mutual Information} (PMI$_{\rm DC}$) rescoring method of
\citet{holtzman-etal-2021-surface} 
to address the surface form competition issue. 
We use $PMI_{\rm DC}=\frac{P(y|x,domain)}{P(y|domain)}$, where $y$ is the candidate and $x$ is the input sentence. 
We observe that $PMI_{\rm DC}$ rescoring only work robustly when all options are reasonable to some extent. 
We filter out low frequency candidate before calculating PMI$_{\rm DC}$ scores. 
An exception to the above is the \textit{location} precondition class, where the answer is biased by context and hints. In this case, we do rescore with $PMI_{\rm DC}$ and simply take the most frequent candidate.

As the LM can sometimes produce outputs that are semantically identical but are only slight lexical variations of each other, we also apply a basic cosine similarity check to filter out these duplicates. We use different thresholds depending on the precondition type. For {\em Item Need} and {\em Location} we use a threshold of $0.75$ due to their shorter generations. For everything else, the default threshold is $0.8$.

Not every event requires all six classes of precondition;
forcing the LM to generate preconditions when they are not needed and will not make sense can result in unpredictable responses. We apply heuristics to determine when a precondition type is unnecessary. First, 
an {\em interaction with others} precondition is only necessary when an event has two peoples' names. Second,
{\em reason} and {\em how} preconditions are necessary only when an event can be expanded syntactically by adding ``because $[X]$'' or ``through $[X]$'', respectively, where $X$ is some sentence continuation.
For example, ``John was arrested \textit{because} John stole a car'' or ``John became a pro wrestler \textit{through} training hard''. We include few-shot examples that contain both positive examples (completion examples that contain keywords), negative examples (completion examples that do not contain keywords) and detect the appearance of keywords in text completion results. 

We do not require {\em reason} preconditions that contain the terms ``want'' or ``need''. 
Reason preconditions that contain those two keywords usually show the intention of the character, e.g., ``John wants to buy a gun'' as a precondition for the action ``John walked to the store''. Unlike {\em reason} preconditions that help to develop the plot by introducing a new event, e.g., ``John's old glass broke" in response to the action ``John bought the new glasses", intentions are generated by looking into the existing part of the story and therefore do not help with expanding the story plot.
Want and need reason-preconditions are deleted automatically.

We require every action to have a \textit{location} precondition. 
Our few-shot examples give examples of responses of ``nothing'', which we treat as a keyword to indicate that an event shouldn't be considered as having a location.
When we detect the nothing response, we delete the precondition.
we use the same strategy with the \textit{item state} precondition.

\subsection{Event Generation}

In POCL-style planning, preconditions are satisfied by the effects of events. 
We directly infer events such that they satisfy a precondition. 
For {\em how}, {\em interaction with others}, and {\em reason} preconditions,
we directly copy the sentence from the precondition to make a new event because the nature of those precondition inferences produce sentences that are also events.
These suggest pairs and chains of events that go together naturally, similar to {\em scripts} \cite{schank_abelson_2008}.
For {\em item need}, {\em location} and {\em item state} preconditions we use specialized prompts for each precondition type:
\begin{itemize}
    \item {\bf To satisfy {\em item need} precondition:} What event occurred so that the character obtained the required item?
    E.g., the {\em item need} precondition ``John has gun'' could be satisfied by ``John bought a gun from the store''.
    \begin{quote}
        Context: $[Sentence]$\\
        How did $[Person]$ get $[item]$\\
        Answer: $[Answer]$
    \end{quote}
\end{itemize}
\begin{itemize}
    \item {\bf To satisfy {\em item state} precondition:} What event need to happen to change the state of an item to its desired state?
    E.g., the {\em item state} precondition ``gun is loaded'' would be true after ``John loaded the gun''.
    \begin{quote}
        Sentence: $[Sentence]$\\
        $[item~state]$
        $[Answer]$
    \end{quote}
\end{itemize}
\begin{itemize}
    \item {\bf To satisfy {\em location} precondition:} What event need to happen for the character to arrive at the desired location?
    E.g., ``John drove to get to John’s house'' causes ``John at John's home'' to be true.
    \begin{quote}
        Context: $[Sentence]$.\\
        How did $[Person]$ get to $[location]$?\\
        Answer: $[Answer]$
    \end{quote}
\end{itemize}

As with precondition generation, each prompt has seven examples and the eighth has the answer blank.
When an event is generated to satisfy a precondition on another event, we create a {\em causal link}, which is a directed arc from the new event to the precondition, indicating that (a)~the precondition is satisfied, and (b)~the new event is a temporal predecessor.
The plot is thus represented as a directed acyclic graph, as shown in Figure~\ref{fig:plan-graph}.

We make a greedy assumption that if more than one event has the same precondition then the same event will satisfy both preconditions. We define two preconditions to be the same if (1)~they are associated with the same character, and (2)~the cosine similarity of the two preconditions surpass certain threshold (0.75 for {\em item need} and {\em location} and 0.8 for other categories).
Thus, before generating a new event, we check to see if an existing event can be used and a causal link extended from the already existing event to the precondition.
This corresponds to the concept of action re-use in POCL planning.
The exception to this rule is when the new causal link causes a cycle in the graph.

When duplicate events are generated, as determined by cosine similarity, we record an event has having multiple possible phrasings. 
The phrasing with the highest count is presented as part of the plan. 
Ties are broken by taking the phrase with the lowest perplexity. Additionally, we also discard the generated precondition if it is identical to its parent event. 
Note, we do not do this for the reverse, where the generated event is identical to the precondition, as shown in Figure 1 in the ``Sally had an argument with John in Sally's house" precondition and event.

Preconditions that match the initial state conditions $I$
are never used to generate events since these are conditions given by the user.

We do not model negative effects of events.
In POCL planning, it is possible for an action to have an effect that negates a precondition.
In that regard, our system is similar to the graph expansion stage of graph plan~\cite{PlotGraph} without the forward-checking phase. We do this for two reasons. First, our planner operates in an open-world and, because preconditions are inferred by a LM, we cannot guarantee we have the complete set of preconditions needed to ensure true logical soundness like in a POCL planner.
Second, stories are relatively permissive of missing events when readers can themselves infer missing details and events. 
For example, if John has an item and then Sam has the item later, the reader doesn't need to be told that John no longer has the item, even though our planner doesn't explicitly represent it.

The open world setting combined with surface form competition introduces the possibility of cycles
where event $e_1$ results in precondition $p_1$ is semantically equivalent to another precondition $p_2$ that is an ancestor ($p_2$ is on a causal path from the goal to to $p_1$).
When this happens, we remove $e_1$ from the plan and backtrack to re-generate that event.

The planner terminates when there are no unsatisfied preconditions.
The algorithm is shown in Algorithm~\ref{alg:pseudoPSO}.

\subsection{Final Total Ordering}

When there are no further preconditions to satisfy, the final step is to produce a total ordering of the events in the plot graph.
In POCL planning, any two events that are not ancestors or descendant of each other are considered partially ordered and any topological sort that doesn't violate any temporal ordering constraints is valid.
Because of our open-world assumptions, we implement a topological sort that prefers certain orderings over others.
Whenever there are two events $e_i, e_j$ that are neither descendants nor ancestors, we order them $e_i > e_j$ when $type(e_i) > type(e_j)$ and $type(\cdot)$ is a function that maps an event to the type of precondition it satisfies and 
\textit{reason} $>$ \textit{how} $>$ \textit{item\_needed} $>$ \textit{item\_state} $>$ \textit{location} $>$ \textit{interaction\_with\_other\_person} $>$ \textit{current action}.
The greater-than symbol means earlier in the totally ordered sequence.
Following cognitive science~\cite{Graesser1991QuestionAI}, events that initiate character intentions should precede other actions that might relate to the pursuit of those intentions.
How things came to be, should also precede events that make use of those establishing events. 
Character interactions, especially conflict,  are often the culmination of many establishing events.

\algnewcommand{\pop}{\textbf{pop}}
\algnewcommand{\genprecond}{generate~preconds}
\algnewcommand{\genevent}{generate~event}
\begin{algorithm}[t]
\scriptsize
\caption{Neural Plot Planner}
\label{alg:pseudoPSO}
\begin{algorithmic}[1]
\State \textbf{Input:} ending event sentence $g$; Initial conditions $I$.
\State Initialize a plan $P\gets \emptyset$; Initialize $queue \gets \{g\}$.
\While{$queue\neq \emptyset$}
    \State Let $event\gets \pop(queue)$ 
    \State Let $context \gets$ sequence of events collected by running a breadth-first search from $event$ to $g$.
    \State Let $\Lambda \gets$ all satisfied preconditions
    \State Let $\Gamma \gets \genprecond(event)$ for each character in $event$ 
    \If {adding any precondition in $\Gamma$ creates a cycle} 
        \State remove $event$ from $P$.
        \State $\Gamma \gets $unsatisfied preconditions due to removing $event$.
    \EndIf
    \For {each $c \in \Gamma$}
        \If {$c \in I$ or $c$ meets conditions for not being expanded} 
            \State $P \gets P \cup \{nil \xrightarrow{c} event\}$ \Comment{Dangling precondition}
        \ElsIf {there exists a precondition $c' \in \Lambda$ that is similar to $c$} 
            \State $event' \gets$ event that satisfies $c'$
            \State $P \gets P \cup \{event' \xrightarrow{c'}
            event\}$ \Comment{Reuse precondition}
        \Else  
            \State $event' \gets \genevent(c, context)$
            \State $P \gets P \cup \{event' \xrightarrow{c} event\}$ \Comment{Satisfy with new event}
            \State $queue \gets queue \cup \{event'\}$
        \EndIf
    \EndFor
\EndWhile
\end{algorithmic}
\end{algorithm}

\section{Evaluation}

We evaluate the extent to which our planning technique generates coherent plots.  
Psychological studies of human reader comprehension~\cite{Graesser1991QuestionAI} measure reader comprehension by reader ability to answer questions about a story.
In our setting, we prompt GPT-3~\cite{gpt-3} to answer {\em enablement} questions about generated plots.\footnote{GPT-3 has been used to evaluate other systems such as summarization~\cite{SummaryEvalGPT3}.} 

We use GPT-3 (text-davinci-002) as an estimate of the ease by which questions can be answered about generated stories, and thereby an estimate of the extent to which generated stories support question-answering for reader comprehension.

We evaluated our system and all baselines by generating plotlines that were seeded from randomly chosen stories from the test split of ROCStories dataset~\cite{ROC}. The ROCStories dataset contains plot-like common everyday stories with titles for each story. 
We randomly sampled 50 story titles from the test split of ROCStories to seed our system and the baselines.
We generate a story plotline for each title and then assess whether the plotline is coherent. Systems are scored by the 
percentage of the time that GPT-3, appropriately prompted, can determine that a given event in a story is enabled by any prior events in the story.

We do not evaluate systems using perplexity or variants of BLEU score.
Perplexity and BLEU compare system outputs to expected outputs in a supervised dataset.
Sentences in a story can be generated that are considered good even if they do not match---or have overlapping tokens---with a ground-truth story. 
Further, our method does not sample from the language model in the conventional way.

\subsection{Coherence Measure}

To measure whether the generated stories support enablement comprehension, we pose questions to GPT-3 in the form ``What's the action, if any, enabled  \textit{ACTION}?" along with the sentences from the earlier parts of the story plan for GPT-3 to choose from. Since the first sentence of a story is not enabled by default, we excluded the first sentence in our evaluation. We set $temperature=0.7$. 

We consider an enablement question \textit{answerable} if the answer is found in the earlier part of the story and is different from (percentage of overlapped 1-gram $\leq0.7$ ) the queried event. If the question yields no answer, GPT-3 will respond with ``none''. 
We include negative few-shot examples in the prompt that are answered with ``none'' so that GPT-3 knows it is a valid response. 
\begin{equation}
score=\frac{\# answerable~enablement~questions}{\# questions}
\end{equation}

To validate our measure, we apply the measure to  100 stories randomly sampled from the test split of the ROCStories dataset.
Stories in the dataset are assumed to be coherent.
In half of the stories, we force incoherence by randomly replacing the queried event with a random even taken from a different story in the dataset.
That is, half of the stories should be answerable  and half of the stories should produce ``none'' when prompted with our above enablement prompt.
Our measure yields an $85.39\%$ response rate on unaltered stories and 
correctly predicts ``none'' $71\%$ of the time when presented with an incoherent story.
Thus our measure has an overall accuracy of $78.20\%$.
ROCStories do not always contain explicitly enabling events. 
It is also sometimes possible for a randomly inserted event to appear to have an enabling predecessor action.
Thus some degree of error is expected, and we deem our measure to be sufficient to produce a relative assessment between plot generation systems that produce story plotlines with similar styles.

\subsection{Models and Baselines}

We evaluated the performance of our neural planner against four baselines:
\begin{itemize}
\item \textit{GPT-J-6B}, which is the same model we use to infer preconditions and events, but allowed to generate an entire story. 
    To induce it to generate plots at a comparable level of abstraction to our system, we prompt it with few-shot  examples pulled randomly from the ROCStories dataset. The prompt used in this baseline can be found in Appendix A.3.
    We set the temperature to 1.2.
    \item \textit{C2PO}~\cite{Ammanabrolu2021AutomatedSV}, which uses bidirectional interpolation using commonsense inference of events. We use the first and last sentence of GPT-J-6B generated stories (see above) in our evaluation set.
    \item \textit{comGen}~\cite{CSKESG}, a transformer based language model fine-tuned on ROCStories and two commonsense knowledge bases. We sampled from the provided stories used in their evaluation to use as a baseline.
    \item \textit{plan-write-revise}~\cite{plan-write-revise} fine-tuned on ROCStories with titles as the topic.
    
\end{itemize}
To ensure we evaluate on only {\em plot}, which are events that change the world state~\cite{dictionary},  
we parse the outputs of the baselines to remove non-action sentences that give declarative statements about fact about the story world and characters, e.g., ``Sam was a high school wrestler.''
We also filtered outputs that consisted of only one sentence, as we cannot fairly compare the logical coherence of a single-event plot against plot lines consisting of multiple events. We provide average generation lengths for each baseline, before and after filtering in Table~\ref{tab:lengths}.

\begin{table}[t]
\centering
\footnotesize
        \begin{tabular}{ccc}
            \toprule
            {\bf System} & {\bf Before filtering} & {\bf After filtering} \\
\hline
            Ours              & 4.16     & 4.08  \\
            C2PO                        & 5.62     & 5.42  \\
            plan write revise           & 5.00     & 4.22  \\ 
            comGen                & 4.98   & 4.35  \\
            GPT-J                       & 3.86     & 3.63  \\
            \hline
            ROC                         & 5.00       & 4.56   \\\bottomrule
        \end{tabular}
        \caption{Average number of sentences for each system before and after pre-processing to remove non-event sentences. This shows that systems are comparable in length when prepared for evaluation.}
        \label{tab:lengths}
\end{table}

Note that none of the baselines we compare to are ending-guided like our system. 
C2PO, the closest, uses bidirectional interpolation. Other baselines are conditioned on leading context. However, our evaluation indicates that the generated plot line's length does not appear to have any significant effects on our evaluation metrics. 

For our system, we require an ending event sentence. We first prompt GPT-J-6B to generate an ending based on the provided title. 
Then, we use the ending to generate a story plot.
Our model
is capable of generating more than one object precondition for a given event.
To make our system more comparable to baselines, which are more linear in generation process, 
we constrain the item-precondition generation to a single precondition.

This is roughly in line with what we observed with the ROCStory dataset, as the majority of events in that dataset tend to involve only one interact-able object.

We didn't evaluate against other story generators such as \citet{ContentA} or \citet{storyGen_via_QA}
because these models tend to generate stories with a lot of dialogue and many descriptive scenes. This creates difficulty when evaluating the logical coherency of those systems' outputs as well as extracting the distinct plot lines to use as a baseline. For consistency of evaluation, we only include methods with outputs that have a similar style and structure as our story plots.

\begin{table}[t]
\centering
\footnotesize
        \begin{tabular}{cc}
            \toprule
            {\bf System}                      &  {\bf enablement(\%)} \\ \cmidrule{1-2}
            Ours         &\textbf{85.71}  \\
            C2PO                        &75.11  \\
            plan write revise           &  67.70  \\ 
            comGen                & 76.65\\
            GPT-J               &71.43  \\
            \hline
            ROCStories         &  85.39  \\
            \bottomrule
        \end{tabular}
        \caption{ The Neural Planner achieves the highest percentage in terms of answerable enablement questions compares to other baselines.}
        \label{tab:results}
\end{table}

\subsection{Results and discussion}

The results are shown in in Table~\ref{tab:results}.
Our planner achieves the highest percentage of answerable enablement questions by a considerable margin. 
This can be partially explained by the fact that our planner is intentionally introducing events to the story that enable future events.
The preconditions and linkages between events are not present in the final rendering of a plotlines and any enablement relations are still implicit. 
Our evaluation metric is likely sensitive to this type of structure.
In that sense, the metric can be seen as a measure of the ease with which enablement questions can be answered about a story.

C2PO and comGen use commonense inferences from COMET~\cite{Bosselut2019COMETCT} that have the potential to create explicit enablement relations. 
C2PO in particular generates commonense inteferences about ``wants'', what is expected to come after an event, and ``needs'', what is expected to precede each event in a story.
As noted by \citet{Ammanabrolu2021AutomatedSV} these commonsense relations between events are very similar to causal links though the exact nature of the relation is implicit.

We repeat the ROCStories response rate of $85.39\%$ in Table~\ref{tab:results} for completeness.
The stories in the ROCStories dataset are coherent but not necessarily written to make answering enablement questions easy to answer---events may be omitted because they are considered obvious. 
Therefore, this comparison is not on a level playing field, and our results do not necessarily imply that our planner is performing above human level.

\subsection{Conclusions}

In this paper, we present a novel 
use of neural language models for generating story plotlines by unifying language models with partial order, causal link planning.
Our system chains backward from a given ending of a story, and reasons about the conditions that are necessary for each event to occur.
It generates events that causally enable subsequent events while backward chaining.
In order to operate in an open world domain and be able to generate stories without predetermined actions schemas or characters, our story planner uses the large, pre-trained language model to infer the conditions and events.

This work suggests that pre-trained language models provide affordances for generating coherent narrative content other than generating continuations from a single prompt. 
Specifically, a large pre-trained language model can operate as a commonsense knowledge base about event preconditions and ways to bring about world conditions when prompted to do so.
While our technique makes heavy use of hand-crafted prompts, all generation is coming from the same model guided by a search algorithm inspired by classical planning.
The ability to construct a search space provides a number of benefits including causal relations between events that correlation with improved coherence and reader comprehension.
It also provides a principled means of reasoning about what sentences should occur in a plot-like story beyond reliance on statistical sampling.

POCL plans are interpretable. One can look at any action in a plan and, through causal links, know why the action is present in the plan. 
Causal links indicate which actions are necessary for other actions to execute, and, more importantly, how each action contributes to the achievement of the goal.
Our technique also inherits a degree of interpretability---despite the use of neural language models---because our plan data structures also contain causal links. 
We cannot guarantee that conditions, and thus links, are missing. 
However, the plan structure generated by our technique provides insights into the decisions that the planner was making during generation.

Our results indicate that story plotlines generated by our planner are coherent as measured by the ability to answer questions about causal enablement relations in generated stories. 
Like POCL planning, generated stories focus on physically grounded action.
One can consider generated plotlines as the ``skeletons''~\cite{simon-spark-2022} for fully fleshed out natural language stories that can include details and dialogue generated later.
This work presents a step forward toward the open research challenge of generating stories in open-world that are also guaranteed to be coherent and comprehensible.


\appendix
\bibliography{aaai23.bib}
\newpage
\section{Appendix A}

\subsection{A.1  Story Samples}
\textit{The stories are generated backwards from the last sentence, which is provided by prompting GPT-J on ROCStory titles.}
{\noindent} \rule[-10pt]{8.5cm}{0.05em}
 John took a train to get to home.\\
 John connects internet to network cable at home\\
 John used the internet to get to internet.\\
 John bought lot of computer game to the internet at internet\\{\noindent} \rule[-10pt]{8.5cm}{0.05em}\\John filled the car with gas on the way to the garage.\\
 John drove to get to garage.\\
 John took home his brown car for a treat at garage\\{\noindent} \rule[-10pt]{8.5cm}{0.05em}\\John walked to get to park.\\
 John buys bird watching lens from shop.\\
 John loves to watch birds at park\\
 John took a taxi to get to bank.\\
 John exchanged the money at the bank.\\
 John went to get to park.\\
 John bought a bird from the pet shop.\\
 John found a lost bird in the park at park\\{\noindent} \rule[-10pt]{8.5cm}{0.05em}\\ John took a bus to get to school.\\
 John buys a pen and paper from school.\\
 John wrote a school paper at school\\{\noindent} \rule[-10pt]{8.5cm}{0.05em}\\John walked to get to office.\\
 John attends a staff meeting at the office.\\
 John attended a staff meeting at office\\{\noindent} \rule[-10pt]{8.5cm}{0.05em}\\ John took a bus to get to school.\\
 John hires party at school\\
 John hold a party in the class at school\\{\noindent} \rule[-10pt]{8.5cm}{0.05em}\\John walked to get to shop.\\
 John buys food at shop.\\
 John managed to consume two more glasses of water at shop\\{\noindent} \rule[-10pt]{8.5cm}{0.05em}\\ John drove to get to police station.\\
 John takes delivery case from the delivery van.\\
 John delivered the case to the police at police station\\{\noindent} \rule[-10pt]{8.5cm}{0.05em}\\John walked to get to computer shop.\\
 John bought a new computer at computer shop\\{\noindent} \rule[-10pt]{8.5cm}{0.05em}\\ John took a taxi to get to school.\\
 John played a prank on his student at school\\{\noindent} \rule[-10pt]{8.5cm}{0.05em}\\John walked to get to shop.\\
 John used his money to buy stuff at shop\\{\noindent} \rule[-10pt]{8.5cm}{0.05em}\\John he got a phone call.\\
 John took a taxi to get to home.\\
 John gets a phone from the table.\\
 Chris took a taxi to get to home.\\
 John he had a call from Chris.\\
 John went to get to department store.\\
 John search for the stolen gifts at department store\\{\noindent} \rule[-10pt]{8.5cm}{0.05em}\\ John walked to get to store.\\
 John buys a banjo from a store.\\
John walked home to get to home.\\
 John played a song on his banjo at home\\{\noindent} \rule[-10pt]{8.5cm}{0.05em}\\ John drove to get to traffic jam.\\
 John drive safely through the traffic jam at traffic jam\\{\noindent} \rule[-10pt]{8.5cm}{0.05em}\\John his patient has heart pain.\\
 John was taken to get to hospital.\\
 John got stethoscope from medical kit.\\
 John diagnosed heart pain in his patient at hospital\\{\noindent} \rule[-10pt]{8.5cm}{0.05em}\\John walked to get to nursery.\\
 John got baby from mother.\\
 John pat the baby in the mother at nursery\\{\noindent} \rule[-10pt]{8.5cm}{0.05em}\\John took a walk to get to bedroom.\\
 John caught bugs in his bed at bedroom\\{\noindent} \rule[-10pt]{8.5cm}{0.05em}\\John gets the car keys from the car.\\
 John drove to get to car.\\
 John fixes car light by car light at car\\
 John check the airbag light in his car at car\\{\noindent} \rule[-10pt]{8.5cm}{0.05em}\\ John drove to get to park.\\
 John played a game with friends at park\\{\noindent} \rule[-10pt]{8.5cm}{0.05em}\\John walked to get to sword shop.\\
 John buys a sword from a sword shop.\\
 John drove to get to park.\\
 John beat his neighbor in a duel at park\\{\noindent} \rule[-10pt]{8.5cm}{0.05em}\\ John took a bus to get to home.\\
 John makes party plan at home\\
 John and his friends plan a party at home\\{\noindent} \rule[-10pt]{8.5cm}{0.05em}\\Liseh walked to get to store.\\
 Liseh gets a phone from a store.\\
 John took a bus to get to office.\\
 Liseh took a bus to get to home.\\
 Liseh gets help from John.\\{\noindent} \rule[-10pt]{8.5cm}{0.05em}\\John loaded the money on the bus.\\
 John took a bus to get to cinema.\\
 John never met his dream girl at cinema\\{\noindent} \rule[-10pt]{8.5cm}{0.05em}\\ John took a bus to get to garden.\\
 John got bitten by a squirrel in his garden at garden\\{\noindent} \rule[-10pt]{8.5cm}{0.05em}\\John climbed down the ladder to get to home.\\
 John gets ladder from the closet.\\
 John took a ladder to get to tower.\\
 John climb to the top of the tower at tower\\{\noindent} \rule[-10pt]{8.5cm}{0.05em}\\ John took a bus to get to restaurant.\\
 John ate some pizza for breakfast at restaurant\\{\noindent} \rule[-10pt]{8.5cm}{0.05em}\\ John took a bus to get to school.\\
 John asked for pencil from teacher.\\
 John asked when he could have the next turn at school\\{\noindent} \rule[-10pt]{8.5cm}{0.05em}\\John get business card from the card board.\\
John walked to get to board room.\\
 John he has to discuss about important business at board room\\
John has booking from hotel service.\\
John took a taxi to get to hotel.\\
 John has meeting room booked from hotel service.\\
 John went to get to board room.\\
 John has to have the meeting at board room\\
 John took a taxi to get to office.\\
 John had a meeting with the boss at office\\
 John went to get to board room.\\
 John gets appointment letter from secretary.\\
 John get board room by appointment at board room\\
John walked to get to cleaning closet.\\
 John gets cleaning cloth from the cleaning closet.\\
 John went to get to board room.\\
 John cleaned the board room at board room\\
 John discuss the important matter in the board room at board room\\{\noindent} \rule[-10pt]{8.5cm}{0.05em}\\John he lost his glasses.\\
 John went to get to optician.\\
 John his old glasses broke at optician\\
 John bought the new glasses at optician\\{\noindent} \rule[-10pt]{8.5cm}{0.05em}\\ John drove to get to office.\\
 John he has to work late at office\\
 John he has to work overtime at office\\
 John has to work late at office\\
John walked home to get to home.\\
 John cannot get a mosquito net at home\\
 John mosquitos chased him at night.\\
 John gets mosquito net from the basket.\\
John ran to get to forest.\\
 John escaped from the mosquitos.\\{\noindent} \rule[-10pt]{8.5cm}{0.05em}\\ John drove to get to car.\\
 John drove his car at car\\{\noindent} \rule[-10pt]{8.5cm}{0.05em}\\John walked to get to beach.\\
 John saw the sunset and the rainbow at beach\\{\noindent} \rule[-10pt]{8.5cm}{0.05em}\\John drove home to get to home.\\
 John pulled the son out of the fire at home\\{\noindent} \rule[-10pt]{8.5cm}{0.05em}\\John walked to get to playground.\\
 John skipped rope for a long time at playground\\{\noindent} \rule[-10pt]{8.5cm}{0.05em}\\ John drove to get to garage.\\
Jack garage wasn't in his neighborhood.\\
 Jack took a cab to get to garage.\\
 John gave jack a check at garage\\
 John gave jack his car service at garage\\{\noindent} \rule[-10pt]{8.5cm}{0.05em}\\John walked to get to kitchen.\\
 John gets a kitchen knife from the kitchen.\\
 John cooks food at kitchen\\
John walked home to get to home.\\
 John he can feed the pugsy at home\\
 John loved pets at home\\
 John fed the pugsy at home\\{\noindent} \rule[-10pt]{8.5cm}{0.05em}\\ John drove to get to team building.\\
 John bought a ticket for team building event at team building\\
 John participate in team building at team building\\{\noindent} \rule[-10pt]{8.5cm}{0.05em}\\ John drove to get to post office.\\
 John didn't feel good about it at post office\\
 John he it didn't like the post at post office\\
 John it made him feel good and enjoy at post office\\
 John enjoyed with the post of amusing mail at post office\\{\noindent} \rule[-10pt]{8.5cm}{0.05em}\\John got flu shot from nurse.\\
 John took a taxi to get to hospital.\\
 John's she has the flu at hospital\\
 John John's girlfriend don't feel well.\\
John walked to get to library.\\
 John lost his girlfriend at library\\
 John bought a book from the library.\\
 John searched for the reason of his broken heart at library\\{\noindent} \rule[-10pt]{8.5cm}{0.05em}\\John walked to get to cinema.\\
 John he didn't like the film at cinema\\
 John watched nothing at cinema\\{\noindent} \rule[-10pt]{8.5cm}{0.05em}\\John walked to get to social media.\\
 John posted rude comments about his friend at social media\\
 John had too much trouble in social media at social media\\{\noindent} \rule[-10pt]{8.5cm}{0.05em}\\John walked to get to cinema.\\
 John buys a movie ticket from ticket office.\\
 John watched the movie at cinema\\{\noindent} \rule[-10pt]{8.5cm}{0.05em}\\John walked to get to ticket office.\\
 John puts the money in the ticket at ticket office\\
John walked to get to cinema.\\
 John buys a ticket from the ticket office.\\
 John walked to get to concert.\\
 John became bored at the concert.\\{\noindent} \rule[-10pt]{8.5cm}{0.05em}\\John his colleagues didn't like him.\\
 John rode a taxi to get to taxi.\\
 John counted the money in a taxi at taxi\\
 John took a taxi to get to office.\\
 John didn't get any raise at office\\
 John doesn't like his colleagues at office\\
 John ordered and finished the slow work day at office\\{\noindent} \rule[-10pt]{8.5cm}{0.05em}\\ John took a bus to get to kitchen.\\
 John gets a spoon from the kitchen.\\
 John swallow the sugar at kitchen\\{\noindent} \rule[-10pt]{8.5cm}{0.05em}\\John walked to get to store.\\
 John buys skates from the store.\\
John walked to get to kitchen.\\
 John sharpened his skates at the kitchen.\\
John walked to get to lake.\\
 John ice skated on the frozen lake at lake\\{\noindent} \rule[-10pt]{8.5cm}{0.05em}\\ John took a taxi to get to party.\\
 Julie drove to get to party.\\
 John and Julie met at a party.\\{\noindent} \rule[-10pt]{8.5cm}{0.05em}\\John walked to get to phone booth.\\
 John buys a phone from the phone booth.\\
 John took a taxi to get to home.\\
 John used the phone to call his friend at home\\{\noindent} \rule[-10pt]{8.5cm}{0.05em}\\John his smoke alarm went off.\\
 John took a car to get to home.\\
 John he feared for his life at home\\
 John fled the smoke alarm at home\\{\noindent} \rule[-10pt]{8.5cm}{0.05em}\\John walked to get to near the river.\\
 John had never had a pet at near the river\\
John walked to get to neighbour.\\
 John bought a puppy from the next door neighbour.
 
 \subsection{A.2. Few-shot Prompt for The GPT-J-6B Baseline}

Here is a story on the topic "Drained Battery".\\
Tom let his friend borrow his phone. The friend kept using it. The friend kept draining the battery. Tom got it back way later. The phone died shortly after.\\

\noindent Here is a story on the topic "Matthew Makes Good And Does Good".\\
Matthew would spend hours working on his pitching skills. By high school, Matthew was being scouted by the big leagues. Matthew became a famous pitcher and had a long career. Matthew's only regret was never having kids. Matthew started a charity to buy sports equipment for poor kids.\\

\noindent Here is a story on the topic "Race".\\
Ray was the slowest on the team. Ray needed a 6 minute mile in order to qualify for the race. Ray decided to train an extra hour everyday. Ray eventually got a 6 minute mile. Ray was able to join the race and he did very well.\\

\noindent Here is a story on the topic "A Constant Struggle".\\
Jeff had always wanted to be a pilot. Jeff would spend hours on flight sims in his garage. Jeff found a school that would teach him how to fly. Jeff signed up to learn to fly a plane that day. Jeff eventually became a pilot for a major airline.

\noindent Here is a story on the topic "$[title]$".\\

\bibliography{aaai23}

\bigskip

\end{document}